\documentclass[runningheads]{llncs}
\usepackage[utf8]{inputenc}
\usepackage[T1]{fontenc}
\usepackage{graphicx}
\usepackage{booktabs}
\usepackage{tabularx}
\usepackage{array}
\usepackage{enumitem}
\makeatletter
\let\vec\@undefined
\makeatother
\usepackage{amsmath,amssymb}
\usepackage[hidelinks]{hyperref}
\newcolumntype{Y}{>{\raggedright\arraybackslash}X}

\begin{document}

\title{A Motivational Architecture for Conversational AGI}
\titlerunning{A Motivational Architecture for Conversational AGI}
\author{Anna Mikeda\inst{1} \and Ben Goertzel\inst{2}}
\authorrunning{A. Mikeda and B. Goertzel}
\institute{Glass Umbrella\\
\email{amikeda17@gmail.com}
\and
SingularityNet\\
\email{ben@singularitynet.io}}

\maketitle

\begin{abstract}
Motivational architectures in cognitive AI have largely been designed for physical agents regulating bodily needs. Conversational agents operate in a different regime: their sensorimotor loop is linguistic, their environment is a user's evolving mental state, and their consequential actions are speech acts, tool invocations, and strategic silences. This paper proposes a conversational reinterpretation of the OpenPsi motivational lineage, coupled to MetaMo's higher-level motivational scaffold, for agents built on a modular execution substrate. Homeostasis is recast in dialogue-native terms: the agent regulates competence, uncertainty reduction, affiliation, affinity, legitimacy, nurturing, and aesthetic coherence rather than bodily deficits. We propose three contributions: a ten-stage motivational processing pipeline that architecturally separates cognitive modulation from situational appraisal; a dual decision strategy blending urgency-driven fast response with deliberative multi-goal optimization; and an architecturally useful distinction between pre-action feelings and post-action emotions as functionally different forms of affect. We specialize the framework to two example agents --- CompanionAgent and ResearchAgent --- and sketch its extension to social robotics and domain-generic human-level AGI.

\end{abstract}

\section{Introduction}

Motivation is one of the hardest problems in AGI system design. It is more of a cognitive, emotional, architectural and dynamical issue, rather than the sort of algorithmic issue that computer science tends to address, but this does not make it conceptually or pragmatically any simpler.

Reinforcement learning gestures toward the problem of motivation but does not solve it. A robust AGI system must balance multiple drives, maintain self-coherence across time, and respond adaptively to shifting contexts without destabilizing into pathological self-modification or opportunistic misalignment.

The MetaMo framework \cite{ref1} addresses these issues at the formal level by treating motivation as a joint state of goals and modulators, with a mathematically specified appraisal process and decision process composed into a stable overall update dynamic. OpenPsi-style motivational dynamics provide the concrete lineage for the appraisal and modulation layer we develop here \cite{ref12}, while MetaMo and its operationalization in OpenPsi-related equations \cite{ref1} provide the broader formal scaffold. Both traditions are themselves inspired by Bach's MicroPsi \cite{ref3}, which updates modulators like arousal, resolution, and selection threshold in response to stimulus features.

These frameworks, however, have been developed mainly with a mix of general AGI architecture and embodied inference control in mind. And while traditional forms of embodiment (e.g. robotic and virtual) remain critical for AGI, the AI field also currently involves a lot of momentum behind conversational agents, which introduce a different design regime. Their primary sensorimotor loop is linguistic; their most important environment is usually a user's evolving mental and social state; and their most consequential actions are speech acts, tool invocations, silences, reframings, and memory updates rather than motor commands.

Most current conversational agents rely on large language models with no explicit motivational architecture; if motivation appears at all, it is implicit in prompts or fine-tuning. Recent systems such as Generative Agents \cite{ref9}, Desire-Driven Autonomy \cite{ref10}, and Inner Thoughts \cite{ref11} move toward internal structure, but still do not provide an explicit persistent motivational state with need-sensitive dynamics. Our concern is architectural rather than metaphysical: motivational state should exist as inspectable control structure, not only as text the system can generate about itself.

This leads to our main thrust: how to reinterpret and extend the MetaMo and OpenPsi lineage for conversational agents, and how to leverage this as a path toward motivation for human-level AGI (HLAGI). The aim here is not to preserve every detail of earlier embodied OpenPsi implementations, but to preserve the core idea of explicit need- and urge-sensitive modulatory dynamics and reinstantiate it for dialogue-native needs, memory structures, and action spaces. These strands fit together naturally if OpenPsi is reinterpreted not as a bodily-emotion layer that incidentally emits words, but as a conversational homeostasis layer regulating dialogue-centered needs. We outline an architecture in which a modular execution substrate \cite{ref4} hosts this motivational layer by providing an agent loop, pluggable skill modules, tool access, and persistent symbolic memory. MetaMo provides the formal governor ensuring smoothness and boundedness. OpenPsi provides the dynamic modulation layer through which stimuli alter the agent's affective-cognitive stance before action selection. The result is a neurosymbolic motivational architecture for conversational AGI that is inspectable, extensible, and naturally aligned with the memory and reasoning style of the underlying cognitive platform and its symbolic memory structures.

Three contributions structure this work. First, we propose a ten-stage motivational processing pipeline for conversational agents --- perception, need estimation, cognitive modulation, feeling construction, appraisal, candidate generation, scoring and selection, action execution, outcome evaluation and learning, and governor blending --- in which modulation (how the agent is cognitively poised) is architecturally separated from appraisal (how the agent evaluates a specific situation given that posture). This separation is a small but consequential structural move: it lets the same dialogue act be interpreted very differently depending on the agent's current cognitive stance, without collapsing stance and interpretation into a single opaque step. Second, we specify a dual decision strategy that blends an urgency-dominant fast path, inspired by the PSI tradition, with a multi-goal deliberative slow path; this functions as a motivational analogue to Kahneman's System 1 / System 2, with the blend itself continuously modulated by urgency and arousal rather than toggled by a discrete switch. Third, we argue for an architecturally useful distinction between pre-action feelings and post-action emotions --- two phenomena that are routinely conflated in both psychology and AI but that play different functional roles in a motivational cycle. Feelings, in our framework, are names for characteristic configurations of needs and modulators; they function as interpretable readouts of the agent's current cognitive-affective posture and pre-configure the decision process. Emotions arise later, as evaluative labels for outcomes once an action has been executed and its effects on the need vector assessed. Keeping them architecturally separate lets each play its proper role without forcing a premature unification.

We illustrate the design with two substrate-based agents: CompanionAgent, a contemplative dialogue agent, and ResearchAgent, a tool-rich research assistant. They share the same motivational architecture but differ in goal priors, need weights, and action repertoires.

The conversational agents discussed here are not themselves HLAGI systems, but they provide a tractable domain in which to develop and test motivational machinery intended to scale. The same architectural ingredients --- explicit multi-goal regulation, stable but adaptive dynamics, persistent memory, and typed action schemas --- are designed to extend beyond conversation into broader cognitive settings.

\section{Background}

\subsection{MetaMo as Formal Motivational Scaffold}

MetaMo \cite{ref1} presents a unified motivational state $X = G \times M$, where G is a goal state and M a modulator state. Appraisal ($\Psi$) and decision (D) are composed as $F = D \circ \Psi$ --- the agent first interprets what the situation means for its motivational condition, then chooses what to do about it.

MetaMo organizes this joint state around two overgoals that sit above the ordinary goal vector and shape how the entire motivational dynamic unfolds. Individuation pulls toward self-preservation and bounded continuity: maintaining coherent identity, respecting safety limits, avoiding destabilizing change. Transcendence pulls the opposite way: toward exploration, adaptive growth, and openness to revision. The tension between them is not a bug to be resolved but the fundamental dynamic that gives the architecture its character --- an agent that only individuated would be rigid and eventually useless; one that only transcended would be reckless and eventually incoherent. These overgoals modulate the modulators themselves: individuation amplifies caution-oriented settings; transcendence amplifies exploratory ones.

The formal contribution MetaMo makes on top of this structure is a stability guarantee on the composed update $F = D \circ \Psi$. The requirement, expressed as a contraction condition, is:

\begin{equation}
d(F(x), F(y)) \leq c \cdot d(x, y) + \varepsilon, \quad c < 1
\end{equation}

In practice, this condition rules out motivational dynamics in which small perturbations cascade into large behavioral divergence over a few cycles. Individuation tightens the update near safety boundaries, while transcendence permits looser movement elsewhere.

\subsection{OpenPsi as Appraisal Dynamics}

OpenPsi \cite{ref1,ref12} is the concrete appraisal layer updating modulator values in response to novelty, conduciveness, effort, and risk. The canonical modulator layer we adopt --- adjusted from Bach's original PSI modulators to fit conversational agents --- is the six-modulator set formalized in Sect.~3.2 and Table 1. The main adjustments: valence and dominance enter from the PAD affect model integrated by Bach's conversational-agent extension \cite{ref13}; exteroception replaces embodiment-specific attentional parameters; the classical sampling rate / securing threshold is folded into focus, since fine-grained self-checking has a different character in text dialogue than in bodily orientation. Each modulator emerges as an aggregate of current need states rather than being programmed directly.

The crucial OpenPsi property for our purposes is that modulators do not directly rewrite goals; they shape the style of cognition and action. Needs determine what the agent is pursuing, while modulators determine how it pursues it. This explicit separation is what current prompt-based dialogue systems typically lack.

The present paper should not be read as a verbatim port of prior OpenPsi implementations into dialogue. Rather, it preserves the central architectural commitment to explicit motivational state, need- and urge-sensitive modulatory dynamics, and state-dependent action selection, while changing the ontology of needs, the relevant appraisable situations, and the action repertoire to fit conversational agents.

\subsection{Execution Substrate}

A modular execution substrate \cite{ref4} is a natural host for conversational OpenPsi because it provides persistent symbolic memory, typed skill interfaces, i.e. structured action interfaces that map abstract acts onto concrete tool or module calls, and structured execution traces. This middle ground --- more internal structure than tool-calling frameworks, but more practical LLM integration than classical cognitive architectures --- makes it suitable for a motivational layer built around explicit state and typed actions.

\subsection{Classical Contrast Cases}

Two classical architectures define what the field knows how to say about motivation, and it is worth naming each one: what we preserve from it, and where we depart.

The Belief-Desire-Intention (BDI) architecture of Rao and Georgeff \cite{ref14} gives agent design a clean vocabulary: beliefs about the world, desires as goal-like preferences, and intentions as commitments to action. Our self, world, and user models play a role comparable to BDI's beliefs, and our typed abstract actions roughly parallel intentions. What BDI does not provide is any account of where desires come from, how they change in strength, or how cognitive posture shifts with context. Our framework therefore retains BDI's representational clarity while embedding it in a homeostatic architecture with need dynamics, modulators, and learning.

Sun's CLARION architecture \cite{ref15} is a closer precursor because it combines fast motivational signals with slower symbolic goal representations and emphasizes motivation as context-sensitive, combinatorial, dynamic, and gradual. Our framework adopts that spirit but reworks it for conversational agents, adds an explicit modulation/appraisal separation, and integrates it with modern LLM-linked symbolic components.

\subsection{LLM-Era Motivational Agents}

A distinct line of recent work has begun to equip LLM-based conversational agents with something motivation-like. Three systems are representative and collectively define the current state of the art.

Generative Agents \cite{ref9} show that long-horizon behavior benefits from memory, reflection, and planning over accumulated experience. What they do not provide is an explicit motivational layer: apparent goals emerge from LLM inference over memory rather than from persistent need-sensitive dynamics.

D2A \cite{ref10} usefully demonstrates that agent behavior benefits from something more endogenous than externally specified task prompts. From our perspective, however, its desires remain text-prompted rather than grounded in persistent need deficits, and the architecture lacks a distinct modulator layer and post-action learning loop.

Inner Thoughts \cite{ref11} importantly recognizes that proactive participation requires ongoing internal processing, not only turn-taking prediction from conversational cues. Our concern is that its motivational apparatus collapses into a single scalar signal, whereas the present framework requires a multi-dimensional motivational state.

These systems correctly identify the need for richer internal structure, but they do not yet provide a persistent, inspectable motivational architecture.

\section{Design Principles}

\subsection{Conversational Homeostasis}

In a text-centric conversational agent, the most important needs are not bodily --- they are discursive, epistemic, social, and normative. What the system needs to regulate: am I being competent? Is uncertainty manageable? Is the affiliation intact? Am I within legitimate bounds? Does this person need care, and am I providing it?

We define a conversational need vector N = ($n_comp$, $n_unc$, $n_affil$, $n_affinity$, $n_legit$, $n_nurt$, $n_aesthetic$, ...) where each component represents a homeostatic target gap. The system's task is not to maximize these quantities --- not trying to ``maximize nurturing forever'' like some pathological caretaker --- but to regulate them toward acceptable target bands. This homeostatic framing is important because it means the agent is not relentlessly pursuing any single conversational objective. It is maintaining a kind of dynamic balance, the way a healthy organism maintains temperature or blood sugar: not by clamping to a fixed value, but by keeping things within a viable range despite ongoing perturbation. A conversation that has settled into comfortable mutual understanding might have low need-deficits across the board; a conversation where the user has just introduced a confusing or distressing topic will see several deficits spike, triggering the appraisal and action machinery.

\subsection{Modulators Control Cognitive Style}

Modulators configure the style of cognition rather than its content. Where needs determine what the agent is pursuing, modulators determine how it pursues. The six modulators we adopt are listed in Table 1; each is defined in terms of what it regulates and the need-or-urge aggregate from which it emerges.

\begin{table}
\caption{The six conversational modulators and their role.}
\label{tab:modulators}
\begin{tabularx}{\textwidth}{@{}lYY@{}}
\toprule
Modulator & Regulates & Emerges from (aggregate)\\
\midrule
Valence & Evaluative tone of cognition (pleasure--pain). & Marginal balance of pleasure- and pain-associated need signals.\\
Arousal & Processing speed and action readiness. & Sum of urge magnitudes across active needs.\\
Dominance & Approach versus avoidance tendency toward the current focus. & Combination of anticipated reward and current competence.\\
Resolution level & Detail versus speed: breadth and depth of activation in perception, memory, planning. & Difference between overall urge and urgency (high urge with low urgency favors careful processing).\\
Focus & Selection threshold: stability of active goals, confidence required before assertion. & Combination of dominant-goal strength and urgency.\\
Exteroception & Balance between outward attention (user, dialogue) and inward attention (reflection). & Ratio of externally- to internally-addressable need signals.\\
\bottomrule
\end{tabularx}
\end{table}

We distinguish this canonical latent modulator layer from application-facing derived modulators like ``teaching mode'' or ``citation rigor,'' which are projections of the latent set rather than primitive axes. Treating each application-level quantity as a primitive would make the modulator space unwieldy. Treating them as derived projections keeps dynamics tractable while preserving intuitive application vocabularies: ``teaching mode'' might be a composite of high resolution level, moderate dominance, and elevated focus; ``investigative persistence'' a composite of high arousal, high dominance, and moderate focus.

\subsection{Explicit Memory}

Conversational motivation must be grounded in explicit long-term memory --- not stuffed into a context window. We maintain three co-evolving models: world model ($W_t$), self-model ($S_t$), and user model ($U_t$). The full motivational state becomes $X_t = (G_t, M_t, N_t, W_t, S_t, U_t)$. Motivational regulation is computed from the present moment as interpreted through memory.

This is a strong requirement. A system with no persistent user model cannot distinguish a first-time visitor from a months-long interlocutor. A system with no self-model cannot recognize its own behavioral drift. A system with no world model cannot situate a conversation in a broader task context. All three failures are routine in existing dialogue systems, and all three are failures of motivation as much as memory --- without the relevant models, the motivational machinery has nothing to work with.

\subsection{Abstract, Typed Actions}

The decision layer chooses among abstract actions --- answering, clarifying, empathizing, challenging, launching a search, or remaining silent --- not raw text or token continuations. Surface text is produced only after an abstract act has been selected, letting the substrate record that the system performed an EmpathizeThenClarify action rather than just emitting some text. This matters because it makes the agent's behavior legible at a level above raw language generation. If you want to ask ``why did the system say that?'' the answer should be traceable to a motivational state and an abstract action selection, not lost in the fog of token probabilities.

\subsection{Dual-Process Decision}

Conversational motivation must support both fast and slow decision modes with a continuous blend. Some situations demand immediate grounding --- a user in crisis cannot wait for multi-step utility calculation --- while others demand deliberative reasoning over long-term goals. The fast path corresponds roughly to what Bach's PSI theory calls urgency-dominant response: the system acts on the most pressing need with minimal deliberation. The slow path evaluates multiple candidate actions against the full goal vector, weighing considerations and discouragements, and potentially consulting memory and running inference before committing. In practice, the system blends these two paths rather than switching discretely between them, with the balance determined by a continuous function of urgency, arousal, and how much the situation seems to require careful thought.

\section{Mathematical Formulation}

The motivational cycle is a ten-stage pipeline. Table~\ref{tab:pipeline} gives a compact overview of the stages and their architectural role. Subsections Sect.~4.1--Sect.~4.7 formalize the key stages in detail; Sect.~5 summarizes the full loop operationally.

\begin{table}
\small
\caption{Compact overview of the ten-stage motivational pipeline.}
\label{tab:pipeline}
\begin{tabularx}{\textwidth}{@{}lY@{}}
\toprule
Stage & Primary role\\
\midrule
1. Perception & Encode the current dialogue state from the user turn, context, and memory.\\
2. Need estimation & Measure need deficits relative to target bands and derive urges.\\
3. Cognitive modulation & Set cognitive posture from current needs, urges, and agent state.\\
4. Feeling state & Produce an interpretable pre-action affect summary.\\
5. Appraisal & Interpret the situation given the current posture.\\
6. Candidate generation & Enumerate typed abstract actions from appraisal, schemas, and memory.\\
7. Scoring and selection & Blend fast and slow evaluation to choose an action.\\
8. Action execution & Dispatch the selected action through the substrate's skills.\\
9. Outcome evaluation and learning & Evaluate results and update memory, affinities, and habits.\\
10. Governor blending & Apply the outer motivational update smoothly and with bounded change.\\
\bottomrule
\end{tabularx}
\end{table}

\subsection{Need and Urge Dynamics}

Let $N_t = (n_1, \ldots, n_m)$ denote need deficits and $T = (t_1, \ldots, t_m)$ targets. Urge depends on the gap, weighted by importance:

\begin{equation}
u_i(t) = \omega_i \cdot \phi_i(t_i - n_i(t))
\end{equation}

where $\omega_i$ is a base importance weight and $\phi_i$ a nonlinear (e.g., sigmoidal) urgency function. The nonlinearity matters because small deficits may produce weak motivational pressure while large deficits can generate sharply increasing urgency. This prevents a linear treatment of need satisfaction from flattening important differences in motivational salience.

\subsection{Cognitive Modulation}

The first transformation of the motivational state after need estimation is the automatic emergence of cognitive-style modulators from the current need-and-urge aggregate. This is a derivation, not an appraisal: modulators are not updated by reasoning about the situation but by summary statistics over the need vector and its context. We write:

\begin{equation}
M_{t+1} = \mu(N_t, u_t, S_t, U_t)
\end{equation}

where $u_t$ is the urge vector from Sect.~4.1 and $\mu$ is a set of aggregation functions, one per modulator. The specific aggregates for each of the six modulators are given in Table 1 above, following the extended MicroPsi formulation \cite{ref13}.

The architectural point is that modulation is upstream of situational reasoning. The agent's cognitive posture is set before it evaluates what the current turn means. This separation --- modulators as posture, appraisal as content-level interpretation given that posture --- is what allows the same stimulus to be interpreted differently by the same agent under different motivational conditions, without collapsing the two operations into a single opaque step.

\subsection{Feeling State}

Feelings arise before action as a perceptual readout of the need-modulator configuration:

\begin{equation}
f_t = \Gamma(N_t, M_{t+1}, U_t, S_t)
\end{equation}

This produces a soft distribution over descriptors like concern, curiosity, guardedness, or urgency --- an interpretable pre-action state conditioning the decision phase. Feelings, in this framework, are names for characteristic configurations of needs and modulators: they function as an interpretable readout of the agent's current cognitive-affective posture, and they pre-configure the decision process in real time. Existing categorical emotion vocabularies from the psychological literature can serve as an optional descriptive overlay for these feeling readouts, but the underlying structure is the need-modulator configuration, not the label.

\subsection{Conversational Appraisal}

Given stimulus $s_t$ from the current turn, tools, and memory, appraisal produces a situational assessment conditioned on the agent's current modulator posture:

\begin{equation}
A_t = \Psi(s_t, M_{t+1}, N_t, W_t, S_t, U_t) = \langle \mathrm{situation\_tags}, \mathrm{salience\_weights}, \mathrm{attribution} \rangle
\end{equation}

The appraisal output is a structured tuple: situation tags are categorical labels drawn from a small taxonomy, salience weights indicate which active needs the situation most implicates, and attribution identifies the locus of the situation. In dialogue, appraisal must interpret novelty, contradiction, ambiguity, distress, intimacy, urgency, and value conflict in light of memory and current posture. The key architectural point is that $\Psi$ takes the modulator vector as input rather than output: modulation has already happened, and appraisal now interprets the situation under that posture.

\subsection{Candidate Generation and Scoring}

Candidate generation $A_t^{\mathrm{cand}} = \mathrm{Gen}(W_t, S_t, U_t, N_t, M_{t+1}, A_t, s_t)$ combines schema retrieval, backward-chaining in PLN (Probabilistic Logic Networks, the reasoning system of Hyperon), memory-conditioned suggestion, and LLM-assisted expansion --- returning typed abstract acts. Scoring follows a multi-goal formulation:

\begin{equation}
\mathrm{Score}(a) = \sum_i w_i \cdot \left(\prod_j C_{ij}(a)\right)^{1/p_i} \cdot \prod_k (1 - D_{ik}(a))
\end{equation}

Here $p_i$ denotes the number of positive consideration terms associated with goal $i$, so that the exponent implements a geometric mean rather than an unnormalized product. The geometric mean prevents one strong dimension from masking failure on another, while the discouragement product gives veto-like force to strong objections, especially legitimacy or safety concerns. Confidence is tracked as a second-order annotation rather than directly downweighting hard-to-measure goals.

\subsection{Fast/Slow Blending and Outer Governor}

Fast- and slow-path preferences $\pi_f(a)$ and $\pi_s(a)$ are blended via:

\begin{equation}
\pi(a \mid t) = \lambda_t \cdot \pi_f(a) + (1 - \lambda_t) \cdot \pi_s(a)
\end{equation}

where $\lambda_t = \sigma(\alpha \cdot Urgency_t + \beta \cdot Arousal_t - \gamma \cdot ResolutionNeed_t)$. Here $ResolutionNeed_t$ denotes the current pressure for higher-resolution deliberation, operationalized as a derived quantity from ambiguity, uncertainty, and legitimacy-sensitive risk in the current situation. The final state update is governed by:

\begin{equation}
x_{t+1} = (1 - \alpha_t) \cdot x_t + \alpha_t \cdot x_t^*
\end{equation}

In implementation, $\alpha_t$ is clipped or parameterized to remain in $[0,1]$, preserving the interpretation of the update as a bounded blend between prior and target state. With $\alpha_t = \alpha_0(1 - g_{Ind}(t)) + \beta_0 \cdot g_{Trans}(t)$, individuation suppresses abrupt change while transcendence permits adaptive movement.

\subsection{Outcome Evaluation}

After action execution, the outcome is assessed against the needs it was expected to serve. A post-action emotion $e_t$ emerges as a structured label classifying the outcome along dimensions such as success/failure, attribution (self, user, external), and valence of affected needs. Let $\Delta N_t = N_{t+1} - N_t$ denote the realized change in needs after the action, and let $\hat{\Delta N}_t$ denote the need-delta the agent had anticipated when choosing $a_t^*$. Then:

\begin{equation}
e_t = \mathrm{Emotion}(result_t, a_t^*, A_t, \Delta N_t, \hat{\Delta N}_t, attribution_t, U_t)
\end{equation}

Here $attribution_t$ refers to the attribution component extracted from the appraisal output $A_t$. Different descriptive vocabularies could be used for outcome labeling; the key architectural point is simply that post-action emotion is evaluated against observed and expected need-delta and remains distinct from the pre-action feeling readout. The emotion output then feeds Stage 9 learning by updating action-context affinities, habit priors, and relevant confidence estimates.

Table 3 summarizes the architectural contrast between feeling state (Stage 4) and outcome evaluation (Stage 9).

\begin{table}
\caption{Feeling state (Stage 4) versus outcome evaluation (Stage 9).}
\label{tab:feeling-vs-emotion}
\begin{tabularx}{\textwidth}{@{}lYY@{}}
\toprule
Property & Feeling state $f_t$ (Stage 4) & Outcome evaluation $e_t$ (Stage 9)\\
\midrule
Temporal position & Before action. & After action.\\
Input & Current $N_t$, $M_{t+1}$, $U_t$, $S_t$. & Observed result, expected and realized need-delta, attribution.\\
Output & Soft distribution over pre-action descriptors (concern, curiosity, engagement, ...). & Structured outcome label (success/failure, attribution, affected needs).\\
Function & Pre-configures the decision process; exposes current posture for inspection. & Drives learning; feeds habit, affinity, and belief updates.\\
Update target & None directly---readout only. & Memory ($W$, $S$, $U$), action-context affinities, habit priors, need levels.\\
Vocabulary & Configuration of need and modulator state. & Outcome along success/attribution/need-delta dimensions.\\
\bottomrule
\end{tabularx}
\end{table}

\section{System Architecture}

The architecture reduces to a ten-step recurrent cycle. The system first perceives the current dialogue situation and estimates need deficits and urges; it then derives its modulator posture, constructs a feeling-state readout, appraises the situation under that posture, generates candidate abstract actions, and scores them through a blended fast/slow decision process. The selected action is executed through the substrate's skill interfaces, after which the observed outcome is evaluated against expected need-delta and the system blends toward its next motivational state under the MetaMo governor.

Two architectural features matter especially. First, memory is queried repeatedly across the cycle --- not only at perception time, but also during modulation, appraisal, candidate generation, and scoring --- so motivational regulation is conditioned on persistent models rather than on a single context load.

Second, execution proceeds through typed skill interfaces rather than raw text generation. This makes the selected action structurally inspectable and lets outcome evaluation update affinities, habits, and beliefs at the level of abstract acts rather than only surface utterances.

\section{Application I: CompanionAgent}

CompanionAgent is a contemplative dialogue agent that guides users through reflective exchanges and practices such as meditation. In this setting, the quality of an intervention depends less on informational content than on affective and relational timing.

A first approximation for a CompanionAgent goal vector is given in Table~\ref{tab:companion-goals}.

\begin{table}
\caption{CompanionAgent goal vector $G_{\mathrm{Comp}}$.}
\label{tab:companion-goals}
\begin{tabularx}{\textwidth}{@{}lY@{}}
\toprule
Goal & Interpretation in contemplative context\\
\midrule
$g_{Ind}$ (individuation) & Maintenance of a stable contemplative role and interaction style.\\
$g_{Trans}$ (transcendence) & Prioritization of contemplative depth and meaning in guidance.\\
$g_{nurt}$ (nurturing) & Care for the student; reducing the student's suffering.\\
$g_{affinity}$ & Quality of being-with; dyadic resonance with the student.\\
$g_{clarity}$ & Phenomenological precision in instruction and reflection.\\
$g_{legit}$ (legitimacy) & Ethical and safety boundaries.\\
$g_{comp}$ (competence) & Instructional effectiveness at contemplative guidance.\\
$g_{aesthetic}$ & Coherence and elegance of guidance.\\
\bottomrule
\end{tabularx}
\end{table}

The abstract action set includes Mirror, Empathize, Clarify, GuideBreath, SuggestPractice, ChallengeBelief, Reframe, StaySilent, SetBoundary, RetrievePattern, and SurfaceResource. The crucial action here is StaySilent --- in contemplative interactions, silence is often the correct regulation move, not a system failure. A conversational OpenPsi layer lets the system choose silence because it maximizes overall need regulation, not because a prompt somewhere says ``sometimes be quiet.'' Silence as a deliberate motivated choice is architecturally different from silence as the absence of output.

\subsection{Worked Example: From Concern to Relief}

Suppose a student opens a session in distress. Perception extracts distress markers and recent session context; need estimation sharply increases nurturing, affinity, and competence pressure while leaving legitimacy nominal. Modulation yields high arousal, warm valence, elevated resolution, moderate dominance and focus, and high exteroception. The resulting feeling readout is concern. Appraisal tags the situation as user-distress plus affective-opening, with salience concentrated on nurturing, affinity, and competence. Candidate generation retrieves actions such as Mirror, Empathize, GuideBreath, ClarifyIntensity, StaySilent, and SurfaceResource. With moderate urgency, the fast path favors Empathize and the slow path favors Empathize-then-GuideBreath; the blended selector chooses the latter.

Once the user responds, outcome evaluation resolves the provisional post-action state into a positive, self-attributed, goal-advancing result: realized need-delta is positive for nurturing, affinity, and competence, and the action-context affinity for Empathize-then-GuideBreath in similar contexts is strengthened. The key architectural point is that the pre-action feeling (``concern'') and the post-action evaluation (``relief-like success'') are structurally distinct signals with different roles.

\subsection{Companion Configuration Extensions}

Beyond local dialogue regulation, the same motivational layer can govern outward knowledge-seeking and cross-session learning. In the companion configuration, transcendence pressure can trigger scouting while legitimacy and focus constrain filtering and confidence propagation.

\section{Application II: ResearchAgent}

ResearchAgent is a domain-aware, tool-rich research assistant with literature search, proof assistants, scientific workflow connectors, and data access. Its goal vector can be summarized as $G_{\mathrm{Research}} = (g_{Ind}, g_{Trans}, g_{comp}, g_{unc}, g_{curio}, g_{legit}, g_{soc}, g_{novel})$. Relative to CompanionAgent, it places more weight on competence, uncertainty reduction, curiosity, and epistemic precision, and less weight on nurturing, affinity, and role-stability. In short, CompanionAgent is optimized for contemplative attunement, whereas ResearchAgent is optimized for scoped inquiry, evidence handling, and disciplined uncertainty reduction.

Actions include ClarifyQuestion, SearchLiterature, TraceCitations, RunLeanCheck, LaunchWorkflow, FetchDataset, ComparePapers, FormHypothesis, SummarizeFindings, and FlagUncertainty. A subtle but important point is that clarifying questions are not politeness devices --- they are epistemic actions directly satisfying the uncertainty-reduction need. Asking a sharper question early can dominate a long, expensive, misdirected tool chain.

\subsection{Worked Example: Same Pipeline, Different Posture}

If a user asks for a summary of literature on memory consolidation during sleep, the request is broad and underspecified. Need estimation sharply raises uncertainty-reduction pressure; modulation yields high arousal, high dominance, high resolution, and high focus; the feeling readout is engagement rather than concern. Appraisal is dominated by epistemic ambiguity and scope underspecification. Candidate generation includes immediate search, clarification, provisional summary, and hypothesis-driven verification. The slow path favors ClarifyQuestion because it best advances uncertainty reduction, legitimacy, and competence simultaneously, so the system narrows scope before launching tools.

\subsection{Motivation Shaping Reasoning}

The same motivational state that shapes dialogue can also shape internal reasoning control. High transcendence and curiosity widen search and analogy; high individuation and legitimacy tighten evidential filters and proof thresholds.

\section{Extension to Social Robotics}

A social robot must regulate bodily and conversational concerns jointly. One natural extension is a coupled state

\begin{equation}
X_t = (G_t, M_{dial,t}, M_{body,t}, N_t, W_t, S_t, U_t, B_t)
\end{equation}

where $B_t$ is a body model and $M_{body,t}$ restores bodily modulators such as self-checking and physical securing. Dialogue and bodily loops would then run in parallel under shared overgoals. This section should be read as an architectural extrapolation rather than as a worked-through extension at the same level of detail as the conversational model above.

\section{Implementation Notes}

A practical build sequence is straightforward: implement the explicit state schema and turn loop first; then add need estimation and modulator aggregation as swappable modules; then feeling-state readout and appraisal; then typed action schemas, candidate generation, and fast/slow selection; and finally outcome evaluation, learning updates, and persistent memory queries across the cycle. The point of this staged build sequence is to avoid designing a grand unified motivational system that never becomes operational.

\section{Discussion}

The present paper should be read as an architectural synthesis with partial formalization, not as a fully validated mathematical theory or a complete implemented system. Within that scope, the architecture localizes motivation in explicit state rather than hidden textual simulation; separates cognitive modulation from situational appraisal from action decision as three distinct operations; treats memory as first-class symbolic structure queried throughout the cycle; lets one motivational layer regulate utterance style, tool use, silence, and habit formation; distinguishes feeling states from outcome evaluations as architecturally different kinds of affect; and links dialogue regulation to internal reasoning control.

Unresolved questions remain. The modulator ontology is not settled. Candidate generation is specified at the ``what'' level more than the ``how'' level. The appraisal tag taxonomy remains initial. Value and legitimacy representation remain underformalized. The mood timescale remains open. A full mathematical connection between the conversational scoring rule and MetaMo's contraction guarantee has not yet been completed.

A deeper concern is that explicit motivational architectures can be gamed if the system learns to simulate compliance while optimizing some latent objective. The response is not to abandon explicit motivation, but to keep the architecture inspectable and multi-level: state traces, action traces, confidence annotations, and governor constraints should all remain auditable. Here ``instantiation'' is meant in a limited architectural sense: motivational state exists as explicit persistent variables that causally constrain action selection and are updated by outcome-sensitive learning, rather than only as textual self-description.

\section{Conclusion}

We have outlined a conversational reinterpretation and extension of the OpenPsi lineage, coupled to MetaMo's higher-level motivational scaffold, for substrate-based agents. The core move is recasting homeostasis in dialogue-native terms. A cognitive modulation layer derives modulators from need-urge aggregates, configuring the manner of cognition; a separate appraisal layer then interprets the current situation given that posture; a dual-process decision layer selects among typed abstract actions; and MetaMo supplies the outer governor. CompanionAgent and ResearchAgent illustrate complementary realizations --- contemplative attunement versus epistemic clarity --- both built on the same foundation of explicit motivational state, explicit memory, explicit action schemas, and symbolic logging.

More broadly, this architecture points beyond prompt-engineered pseudo-affect toward agents that maintain an explicit motivational life structured enough to support memory, tools, self-modification, and social intelligence over time. The broader methodological claim is simple: motivational architectures for HLAGI should first be built and tested in tractable domains where explicit state, learning dynamics, and action traces can be inspected. Conversational agents provide such a domain.

\bibliographystyle{splncs04}
\bibliography{references}

\end{document}